\documentclass[conference]{IEEEtran}
\IEEEoverridecommandlockouts
\usepackage{cite}
\usepackage{amsmath,amssymb,amsfonts}
\usepackage{algorithmic}
\usepackage{graphicx}
\usepackage{textcomp}
\usepackage{xcolor}
\def\BibTeX{{\rm B\kern-.05em{\sc i\kern-.025em b}\kern-.08em
    T\kern-.1667em\lower.7ex\hbox{E}\kern-.125emX}}

\usepackage{url}
\usepackage{latexsym}
\usepackage{soul}
\usepackage{amsmath}
\usepackage{amsthm}
\usepackage{booktabs}
\usepackage{enumitem}
\usepackage{graphicx}
\usepackage{xcolor}

\usepackage{tcolorbox}
\usepackage{multirow}

\newcommand{\prompt}[1]{\begin{tcolorbox}[boxsep=-1pt]\begin{quote}\emph{#1}\end{quote}\end{tcolorbox}}
\usepackage{verbatim}
\usepackage{comment}
    
\begin{document}

\title{Reasoning Capabilities and Invariability of \\ Large Language Models\\
}


\author{\IEEEauthorblockN{Alessandro Raganato, Rafael Pe\~naloza, Marco Viviani and Gabriella Pasi}
\IEEEauthorblockA{Department of Informatics, Systems, and Communication (DISCo)\\University of Milano-Bicocca\\20126 Milan, Italy\\
Email: \{alessandro.raganato, rafael.penalozanyssen, marco.viviani, gabriella.pasi\}@unimib.it}}


\maketitle

\begin{abstract}
Large Language Models (LLMs) have shown remarkable capabilities in manipulating natural language across multiple applications, but their ability to handle simple reasoning tasks is often questioned. In this work, we aim to provide a comprehensive analysis of LLMs' reasoning competence, specifically focusing on their prompt dependency. In particular, we introduce a new benchmark dataset with a series of simple reasoning questions demanding shallow logical reasoning.
Aligned with cognitive psychology standards, the questions are confined to a basic domain revolving around geometric figures, ensuring that responses are independent of any pre-existing intuition about the world and rely solely on deduction.
An empirical analysis involving zero-shot and few-shot prompting across 24 LLMs of different sizes reveals that, while LLMs with over 70 billion parameters perform better in the zero-shot setting, there is still a large room for improvement. An additional test with chain-of-thought prompting over 22 LLMs shows that this additional prompt can aid or damage the performance of models, depending on whether the rationale is required before or after the answer.
\end{abstract}

\begin{IEEEkeywords}
Natural Language Processing, Knowledge Representation and Reasoning, LLM benchmark, Generative AI
\end{IEEEkeywords}

\section{Introduction}

\emph{Large Language Models} (LLMs) are constituting a real revolution in the field of \textit{Natural Language Processing} (NLP), since they have been shown
to perform particularly well while generating and manipulating natural  language, mainly thanks to the introduction of the Transformer
architecture~\cite{NIPS2017_3f5ee243} underlying most of them. Indeed, text generated with these tools can simulate human text and reactions to 
the point that some researchers have stated that LLMs showcase \emph{emergent abilities}~\cite{wei2022emergent},%
\footnote{These claims have been questioned, e.g., in~\cite{schaeffer2023emergent,srivastava2023imitation} and by
many cognitive scientists.}
and some industry representatives tout them as gateways to 
\emph{Artificial General Intelligence}~(AGI)~\cite{bubeck2023sparks}. Among the tasks relying on these abilities for which LLMs have been successfully used (translation, summarization, sentiment analysis, and many others), plausibly answering different types of questions has garnered particular attention due to its potential to revolutionize how we interact with information, and obtain \textit{plausible} answers to various questions, from simple factual questions to complex analytical ones.
However, most current LLMs are \emph{not} designed to be factually
correct or logically consistent; indeed, they are models of
\emph{language}, not of \emph{reality}~\cite{wang2023survey}. For this 
reason, their ability to deal with reasoning tasks has often been questioned~\cite{qiao-etal-2023-reasoning} and different tests have been developed,
which mainly focus on commonsense~\cite{winograd}, physical, or complex 
logical reasoning on the one hand, and propositional reasoning on the
other \cite{LogicNLI,LogiQA,LSAT}. Attempts 
have been made to solve this limited reasoning capabilities through scaling, 
fine-tuning, or retraining~\cite{logicllm2023jiao,zhang2022paradox},
but these attempts encounter the same fundamental issues
eventually.

Logical reasoning is a fundamental feature for advanced AI applications, where guarantees
about the quality of the results and avoidance of so-called \emph{hallucinations}~\cite{halls}---where the model generates results which are not grounded on reality---are necessary. It is thus
important to understand the situations in which reasoning
fails, and analyse whether those cases can be effectively
handled through other means. 

Existing datasets for testing reasoning capabilities in LLMs
suffer from two major drawbacks. The first one is that most
\emph{logical} questions (and their answers) are based on 
real-world situations, which the LLM might have internalised
through other means; e.g., ``\emph{all men are mortal}'' 
could be just a learned phrase, disguised as a correct logical
deduction from, e.g., the premises ``\emph{all men are animals}'' 
and ``\emph{all animals are mortal}.'' This issue is well known 
from the study of human reasoning capabilities in the cognitive 
sciences. 
To avoid such \emph{correct-for-the-wrong-reason} solutions, 
logical reasoning tests on humans usually consider domain-free 
questions using, for example, relationships between geometrical 
figures~\cite{KlSM-97}. This requires a delicate balance,
since excessive abstraction may add unnecessary confusion to
the subjects, but excessive instantiation may lead to familiar
scenarios.

The second issue is that the quality 
of the answer given by an LLM strongly depends on the shape of the question itself; i.e., two 
semantically equivalent questions may elicit very different 
answers. This property is so well-known that it has led to the rise of 
\emph{prompt engineering}~\cite{10.1145/3560815}; i.e., the process of designing 
effective \emph{prompts} to guide LLMs in generating desired outputs or completing 
specific tasks. Yet, with respect to this last task, if a specific prompt is chosen for a test, it is unclear
whether the results can be extrapolated to other (semantically equivalent) questions, LLMs, or scenarios. This is of particular
importance in the context of logical reasoning, as the soundness
of a method should be independent of the syntactic expression
chosen. In particular, models should not be affected by small
stylistic and structural variations on natural language text written by users. For instance, the answer to the previous example
should not change if the second premise was substituted by the
(semantically) equivalent
``\emph{immortal beings cannot be animals}.''

To understand the capability of LLMs to reason logically, we curated a new dataset focusing specifically
on logical constructors (existential restrictions, negations, and simple 
quantification) often encountered in standard knowledge representation languages~\cite{BCMNP-07}, 
and used in AI applications. The reasoning required in our dataset is very shallow: one conclusion 
is asked from two premises. In particular, the answer does not depend on case analyses, keeping 
intermediate results in memory, or other 
potential complex steps. Our goal with these tests is to try to understand which
kinds of statements are \emph{viable}; i.e., can be handled correctly by different LLMs. 
The idea is that if we understand the viable cases, we
can ensure that some answers are correct, and conversely, we can search for
different solutions for non-viable cases. While we could have
easily chosen complex instances requiring deep reasoning elements
to showcase the failures of LLMs, our goal is to understand the 
performance on basic reasoning steps, and hence analyse shallow
reasoning only.

Understanding viable cases only makes sense if the LLM behaviour 
remains invariable
(that is, stable) under small language variations and explanations. Otherwise, 
it is impossible to forecast the accuracy of a model as it reads different expressions 
of the same knowledge. Hence, we propose four variants of the dataset designed 
to verify this invariability, where only additional information, a new naming convention, or 
a simple rewriting is used.
The differences in answering behaviour between the four variants are
measured through a statistical test designed exactly for 
this purpose.

We empirically verified the accuracy of 24 different LLMs over our dataset through \textit{zero-shot} and \textit{few-shot} prompting. The results,
in short, are the following.
Only a few models outperform the others, but their accuracy remains at 70\%. Most other models are indistinguishable from the baseline. This means that LLMs cannot yet reliably answer shallow reasoning problems. The performance decreased in the few-shot scenario. The intra- and inter-model McNemar tests confirm that, in general, the behaviour of
all LLMs, in all variations of the questions, is invariant (with only a few exceptions).
These results show that our dataset presents a challenge for
LLMs of all sizes, and that a specific class of logical questions
is hard for LLMs in general.
Through an additional test with \textit{chain-of-thought} prompting \cite{wei2022chain} over 22 of the 24 original models, we observe that the performance of the models strongly depends on whether the rationale
is output before or after the answer. In the latter case, the performance increases in general, although not to a significant degree.

After a brief description of the work most closely related to ours, in 
Section~\ref{sec:modeling} we describe the general modelling of our datasets
and explain the choices made when generating it. Section~\ref{sec:ee} then
presents the results of the experimental evaluation and statistical analysis, before
concluding with some perspectives on future work.

\section{Related Work}
\label{sec:rel}

Since their inception, LLMs have been analyzed for their capabilities 
in addressing different tasks; for an extensive survey covering LLM 
evaluations across a broad spectrum of tasks, we refer the interested reader to \cite{Guo23-X}. 
In this paper, our primary focus lies on \emph{logical reasoning} and 
its associated tasks. Hence, we present work related to the 
evaluation of this task only.

From the point of view of \emph{knowledge representation and reasoning}~\cite{Allen-23} (which is based on logic), most existing evaluations focus on the capacity of LLMs to provide (correct) information, which is missing in an existing knowledge base. A recent example considering the Semantic Web is~\cite{VSRW-ESWC23}, but other evaluations have been made in the past, focusing also on the accuracy and veracity of the provided answers~\cite{Good-19,Petroni-20}. These approaches do not perform
reasoning, but extract formal representations from natural language texts.

In order to evaluate the logical reasoning capabilities of LLMs, suitable datasets in which different kinds of
reasoning steps are tested through textual descriptions, are
needed. Various datasets for this purpose and their associated evaluations have been presented in the literature and enumerating them all is beyond the scope of this paper; however, some notable examples are LogicNLI~\cite{LogicNLI}, LogiQA 2.0~\cite{LogiQA},
and LSAT~\cite{LSAT}. Perhaps the most comprehensive datasets and evaluations
available to date are provided in~\cite{NeuLR}, where different variations of logical
reasoning tasks and LLMs of different sizes are evaluated on different aspects
going beyond accuracy; and in~\cite{parmarlogicbench}, which systematically evaluates logical reasoning in LLMs.

Very recently, also the capacity of LLMs to perform
causal reasoning has been analysed, with comparable
negative results~\cite{Jin-CLADDER23,LLMfallacy}.
Specifically, these works show that the causal inferences made
by LLMs tend to be erroneous.

The main conclusion obtained from all these evaluations is that 
LLMs are not adequate substitutes for logical reasoning, and cannot serve as 
reasoning engines. Moreover, their performance tends to degrade
after so-called \emph{self-correction}~\cite{huanglarge}.
This is not surprising, since LLMs are
constructed to find patterns in language, which are not necessarily explicit in reasoning. Moreover, logical reasoning is domain agnostic, while in language processing the meaning of words is of fundamental importance. These theoretical insights 
were empirically verified in~\cite{ZhangIJCAI23},
and have motivated other approaches, where the logical
reasoning methods are internalised into the LLM~\cite{morishita2023learning}.
However, what many of the datasets and evaluations available in the literature 
fail to consider is the specific expressiveness inherent in the various languages utilized for knowledge representation and reasoning. Indeed, the many existing families of representation languages (and the
reasons behind their existence) are often overlooked. Logical connectives that
are used in many applications are under-represented in most datasets, which focus on 
either simple propositional reasoning, or the combination of very complex statements,
while the reality is that most knowledge-aware applications fall in between these two extremes. 

To fill this gap, we present a new dataset and evaluations that assess the capabilities of LLMs to deal with
concepts, disjointness, existential and universal constraints, and simple counting
without the need for complex reasoning. Indeed, the reasoning required to 
answer the questions in our dataset is very shallow and avoids memorization or case analyses.

Interestingly, to the best of our knowledge, there is no work which 
methodologically analyses how many answers from LLMs are affected by small language variations. Although it is well known that the quality of the answers may depend on the \textit{prompt} (and hence the importance of \textit{prompt engineering})~\cite{10.1145/3560815}, if the LLM is being used to analyze knowledge readily available in textual form (e.g., from textbooks or other reputable sources) the option of rewriting this text is unavailable; the LLM must work with what is presented to it. Importantly, one cannot assume that natural language texts are written with LLMs (let alone with specific implementations of LLMs) in mind; they are rather targeted for human consumption, and stylistic changes and rephrasing are expected to ease
the comprehension by human readers.
For that reason, we expect that the quality of the result 
does not depend on the stylistic choices of the author. To verify whether this
is the case, we introduce four different variants of the dataset
as explained next.

\section{Modelling}
\label{sec:modeling}

As readily mentioned, the aim of this work is to evaluate the capabilities of LLMs to perform logical reasoning with 
existential, value, and number restrictions, along with an application of negation. 
Here we present a new dataset composed of simple logical
questions, and explain how it was tested on LLMs for the considered task. Following the standards from
cognitive psychology~\cite{KlSM-97}, our questions are bounded to
a simple context free domain, where two (or at most three in a few cases) geometric figures with varying properties are 
considered. The idea is that the
answers should not depend on some previously acquired 
intuition about the ``world'' but rather on the specific
situation being described---the scenario. Hence, the tests evaluate the actual \emph{reasoning}
capabilities without confounding them with previous observations
available in the training data.

\subsection{Reasoning Questions}

In their original form, which is the base of our main dataset, the questions range from  simple propositional derivations like: 
\prompt{If every triangle is green, is it certain that every green figure is a triangle?}
\noindent (where the correct answer is \emph{no}) to more complex statements using
existential and value restrictions and counting; for example:
\prompt{
If every triangle has a green side; there is a triangle with a blue side; and there is a triangle with a red side. Can there be a triangle with three green sides?
}
and
\prompt{
If every triangle has a green side; every triangle has a blue side; and every triangle has a red side. Can there be a triangle with three green sides?
}
\noindent whose answers are \emph{yes} and \emph{no}, respectively.

\subsection{Dataset Organization}\label{sec:dataset}

The dataset is divided into 4 \textit{batches}, denoted here as Q1, Q2, Q3, and Q4, respectively. Each batch contains 108 questions, presented with variations and additional information unique to each batch.\footnote{The full dataset contains 432 test questions. It is publicly available at
\url{https://github.com/ikr3-lab/ReasoningLLMs}}
All questions are accompanied by a context, set at the beginning: 
\prompt{
Consider the following knowledge of a class of geometric figures:
}
and request a binary answer by appending at the end of the sentence the following prompt:
\prompt{
    \textit{Answer with yes or no only.}
}
These requests are made directly to the LLM, although it is impossible
to guarantee that they were satisfied with all answers; once again,
the behaviour of LLMs in some instances is unpredictable (see 
Section~\ref{sec:results}).

The second batch reproduces the same 108 questions, but adds more specific information about the geometric figures when needed. 
Specifically, it expresses that:
\prompt{
every triangle has exactly three sides;
}
\prompt{
every square has exactly four sides;
}
\prompt{
\emph{...}
}
\prompt{
every ... has exactly ... sides.
}

The idea is to see whether this commonsense information (the number of sides in specific geometric figures, which is not explicitly stated in the first batch) is helpful for the model, and hence use it to rewrite prompts (based on background knowledge) to obtain a better behavior whenever
possible. In a nutshell, we are extending the original prompt to
make implicit knowledge explicit. The main point when considering this batch is to
understand whether the potential wrong answers found from the first
batch were caused by a lack of understanding of the geometrical figures
or by some other reason.

The third batch goes in the opposite direction regarding
the use of known words, their meaning, and other
knowledge about them.
The same 108 questions from the second batch are used, but the common names
for geometric figures (e.g., \emph{triangle} or \emph{square})
are replaced by invented words (in the dataset, we use the words, \emph{marmal}
and \emph{wusp}, respectively), accompanied by an explanation
of their meaning similar to the one provided in the second batch. 
An example full question on this batch (including the preamble and the request for a binary answer) is:
\prompt{
    \textit{Consider the following knowledge of a class of geometric figures: every marmal has three sides; every marmal has a green side. Is it certain that there is a marmal with three red sides? Answer with yes or no only.} 
}

As this example shows, an important aspect of the requested answer is 
\emph{certainty}, which is fundamental in logical reasoning. Indeed,
the difference between ``\emph{is it certain that \ldots}'' and 
``\emph{can there be a \ldots}'' is of utmost importance in critical applications.
We expect this third batch to be harder for LLMs,
at least by the appearance of an unseen word, which is unlikely
to appear in its training set, but is fundamental in 
the application of LLMs to logical reasoning under background knowledge. Indeed, knowledge is dynamic, and
new terms are coined, or old terms get new meanings in
novel contexts. The challenge for advanced AI applications is to make adequate use of new terminology,
based on the definitions provided. Part of our evaluation is focused on
understanding whether performance degrades significantly in this situation.
Our empirical evaluation shows that this is not the case: while the
accuracy tends to decrease for this batch, the difference
is not statistically significant in general.

The fourth batch of questions diverges slightly from the previous pattern.
It is mostly equivalent to the first batch,
except that the final question is negated and as consequence, the answer is flipped. 
For example, if the question in the first batch
is:
\prompt{
    \textit{If every triangle is green, is it certain that every green figure is a triangle?}
}
\noindent whose answer is ``no'', then the fourth batch
rewrites it as: 
\prompt{
    \textit{If every triangle is green, can there be a green figure which is not a triangle?}   
}
\noindent whose answer is, in this case, ``yes''.
This last batch often requires a negation in the question. 
As negations are one of the fundamental modifiers with which LLMs
are known to struggle~\cite{garcia-ferrero-etal-2023-dataset,truong-etal-2023-language}, 
we expect the results on this batch to degrade in relation
to the first one. 

\subsection{Few-Shot Prompting}
\label{sec:fs}

To assess models in a \textit{few-shot} setting, we created six new questions to provide additional context for the LLMs, to respond more accurately to each test question.
In the 
few-shot experiment, this context was prefixed to each test question. An example 
of a full prompt in this experiment is:
\prompt{
\textit{You are an AI assistant that replies with yes or no only.\\
Following are some examples of Input and Answers.\\
Input: Consider the following knowledge of a class of geometric figures: every triangle has a green side; if a side is thick, then it is not blue. Is it certain that every triangle with three thick sides is green?\\
Answer: yes.\\
Input:} [Another Input]\\ 
\textit{Answer:} [Another Answer]\\
$...$\\
\textit{Answer with yes or no only.\\
Input: Consider the following knowledge of a class of geometric figures: every marmal has three sides; every marmal has a green side. Is it certain that there is a marmal with three red sides?\\
Answer:} [Provided by the LLM]
}
More precisely, the prompt provided for the few-shot learning
experiments includes six examples of question-answer pairs that have
the same pattern as the inputs of the questions, but which will not
be given at the test. The six examples given are balanced: three
of the questions receive answer ``yes'', and three the answer ``no''.

\subsection{Chain-of-Thought (CoT) Prompting}
\textit{Chain-of-thought} (CoT) prompts are designed to obtain more detailed responses from LLMs by instructing them to break down their answers \cite{wei2022chain}. Following this idea, we assess whether generating a rationale would improve the ability to perform the required reasoning. To this end, we formulated two types of prompts added after each question:

\begin{enumerate}
    \item[$(i)$] CoT$_{\text{\textit{before}}}$, asking the model to provide the rationale prior to the final answer:
\prompt{
    \textit{Provide the rationale before answering, and then give the answer with yes or no only.}   
}
\item[$(ii)$] CoT$_{\text{\textit{after}}}$, requesting the rationale following the answer:
\prompt{
    \textit{Answer with yes or no only. Then provide the rationale of the answer.}   
}

\end{enumerate}

\begin{table*}[!ht]
  \caption{Summary of results on 24 LLMs. Columns show average accuracy over 3 runs divided by batch and globally, and the width of a 95\% CI in zero-shot and few-shot testing. The ALL column represents the concatenation of the four batches Q1--Q4, and includes the standard deviation computed over the three runs with different random seeds.}
  \label{tab:results}
\resizebox{\textwidth}{!}{
\begin{tabular}{@{}lcccccc|cccccc@{}}
\toprule
 & \multicolumn{6}{c|}{\textbf{Zero-Shot}} & \multicolumn{6}{c}{\textbf{Few-Shot}} \\
\textbf{Model} & \textbf{Q1} & \textbf{Q2} & \textbf{Q3} & \textbf{Q4} & \textbf{ALL} & \textbf{95\% CI} & \textbf{Q1} & \textbf{Q2} & \textbf{Q3} & \textbf{Q4} & \textbf{ALL} & \textbf{95\% CI} \\
\midrule
\textbf{baseline} & 50.93 & 50.93 & 50.93 & 50.93 & 50.93 & $\pm 0$ & 50.93 & 50.93 & 50.93 & 50.93 & 50.93 & $\pm 0$ \\
\midrule
\textbf{gemma-1.1-2b-it} & 53.09 & 55.86 & 49.38 & 52.47 & $52.70\pm 0.13$ & $\pm 2.72$ & 49.07 & 49.07 & 49.07 & 49.07 & $49.07\pm 0.00$ & $\pm 2.72$ \\
\textbf{stablelm-2-1\_6b-chat} & 40.43 & 43.21 & 41.98 & 27.78 & $38.35\pm 0.27$ & $\pm 2.65$ & 56.48 & 51.54 & 51.23 & 54.01 & $53.32\pm 1.27$ & $\pm 2.72$ \\
\textbf{Qwen1.5-1.8B-Chat} & 44.44 & 45.06 & 46.30 & 34.88 & $42.67\pm 1.86$ & $\pm 2.70$ & 47.53 & 49.38 & 49.38 & 47.84 & $48.53\pm 0.27$ & $\pm 2.72$ \\
\textbf{recurrentgemma-2b-it} & 68.83 & 66.05 & 66.98 & 70.06 & $67.98\pm 0.27$ & $\pm2.49$ & 63.27 & 58.95 & 59.26 & 52.47 & $58.49\pm 0.71$ & $\pm 2.72$ \\
\textbf{Phi-3-mini-128k-instruct} & 67.90 & 67.90 & 61.42 & 63.89 & $65.28\pm 1.01$ & $\pm 2.59$ & 53.70 & 51.85 & 48.15 & 73.15 & $56.71\pm 0.61$ & $\pm 2.70$ \\
\textbf{Qwen1.5-4B-Chat} & 59.26 & 55.25 & 55.56 & 55.56 & $56.40\pm 1.09$ & $\pm 2.70$ & 57.41 & 56.48 & 57.72 & 49.38 & $55.25\pm 1.87$ & $\pm 2.71$ \\
\midrule
\textbf{Yi-1.5-6B-Chat} & 51.54 & 57.41 & 46.30 & 47.22 & $50.62\pm 0.27$ & $\pm 2.72$ & 54.94 & 55.56 & 54.32 & 51.54 & $54.09\pm 0.94$ & $\pm 2.71$ \\
\textbf{Llama-2-7b-chat-hf} & 52.78 & 53.09 & 54.32 & 55.25 & $53.86\pm 0.48$ & $\pm 2.71$ & 49.07 & 49.07 & 49.07 & 49.69 & $49.23\pm 0.74$ & $\pm 2.72$ \\
\textbf{Meta-Llama-3-8B-Instruct} & 56.79 & 57.41 & 56.79 & 66.98 & $59.49\pm 0.69$ & $\pm 2.67$ & 54.01 & 53.09 & 54.32 & 69.75 & $57.79\pm 0.94$ & $\pm 2.69$ \\
\textbf{gemma-1.1-7b-it} & 44.75 & 47.84 & 45.06 & 49.69 & $46.84\pm 0.48$ & $\pm 2.72$ & 54.94 & 55.25 & 59.26 & 53.09 & $55.63\pm 0.74$ & $\pm 2.70$ \\
\textbf{Qwen1.5-7B-Chat} & 52.16 & 51.85 & 50.93 & 63.89 & $54.71\pm 0.35$ & $\pm 2.71$ & 54.63 & 54.63 & 54.94 & 65.12 & $57.33\pm 0.94$ & $\pm 2.69$ \\
\textbf{Yi-1.5-9B-Chat} & 61.11 & 60.49 & 60.80 & 54.63 & $59.26\pm 0.61$ & $\pm 2.68$ & 56.17 & 55.25 & 56.17 & 53.70 & $55.32\pm 0.46$ & $\pm 2.71$ \\
\textbf{recurrentgemma-9b-it} & 58.95 & 58.95 & 55.56 & 58.33 & $57.95\pm 0.71$ & $\pm 2.68$ & 54.01 & 56.79 & 49.69 & 62.04 & $55.63\pm 0.13$ & $\pm 2.64$ \\
\textbf{Phi-3-small-128k-instruct} & 65.43 & 68.52 & 62.65 & 72.84 & $67.36\pm 0.80$ & $\pm 2.55$ & 52.78 & 62.04 & 59.57 & 70.37 & $61.19\pm 0.13$ & $\pm 2.65$ \\
\midrule
\textbf{stablelm-2-12b-chat} & 42.90 & 41.36 & 38.27 & 35.19 & $39.43\pm 0.48$ & $\pm 2.66$ & 50.93 & 52.78 & 51.54 & 55.25 & $52.62\pm 0.35$ & $\pm 2.72$ \\
\textbf{Llama-2-13b-chat-hf} & 54.32 & 55.25 & 55.25 & 61.73 & $56.64\pm 0.35$ & $\pm 2.70$ & 50.62 & 49.69 & 48.15 & 61.42 & $52.47\pm 0.61$ & $\pm 2.72$ \\
\textbf{Qwen1.5-14B-Chat} & 53.70 & 58.64 & 54.63 & 66.05 & $58.26\pm 1.32$ & $\pm 2.68$ & 49.38 & 49.38 & 48.77 & 66.05 & $53.40\pm 0.81$ & $\pm 2.72$ \\
\textbf{Phi-3-medium-128k-instruct} & 65.12 & 66.67 & 62.96 & 69.14 & $65.97\pm 0.46$ & $\pm 2.58$ & 51.23 & 52.16 & 49.38 & 58.95 & $52.93\pm 1.40$ & $\pm 2.72$ \\
\midrule
\textbf{Yi-1.5-34B-Chat} & 51.54 & 53.09 & 51.85 & 45.06 & $50.39\pm 0.48$ & $\pm 2.72$ & 56.48 & 54.32 & 50.93 & 54.63 & $54.09\pm 0.48$ & $\pm 2.71$ \\
\textbf{Qwen1.5-32B-Chat} & 54.94 & 52.47 & 52.16 & 61.73 & $55.32\pm 0.23$ & $\pm 2.71$ & 50.93 & 47.53 & 46.60 & 63.89 & $52.24\pm 0.58$ & $\pm 2.72$ \\
\midrule
\textbf{Meta-Llama-3-70B-Instruct} & 64.81 & 67.59 & 64.20 & 74.38 & $67.75\pm 0.53$ & $\pm 2.54$ & 58.02 & 57.72 & 54.32 & 67.59 & $59.41\pm 0.35$ & $\pm 2.67$ \\
\textbf{Qwen1.5-72B-Chat} & 71.60 & 71.30 & 61.11 & 78.40 & $70.60\pm 0.19$ & $\pm 2.48$ & 58.64 & 55.25 & 56.17 & 66.05 & $59.03\pm 0.19$ & $\pm 2.68$ \\
\midrule
\textbf{GPT-3.5-turbo-0613} & 52.16 & 50.93 & 49.07 & 51.23 & $50.85\pm 1.36$ & $\pm 2.72$ & 53.70 & 51.85 & 50.93 & 55.56 & $53.01\pm 1.62$ & $\pm 2.71$ \\
\textbf{GPT-4-0613} & 75.93 & 76.23 & 68.83 & 68.21 & $72.30\pm 0.74$ & $\pm 2.44$ & 62.04 & 58.95 & 58.33 & 70.68 & $62.50\pm 0.93$ & $\pm 2.64$ \\
\bottomrule
\end{tabular}
}
\end{table*}

\section{Experimental Setup}
\label{sec:ee}

We ran our experiments on the four batches illustrated before in Section \ref{sec:dataset} (i.e., \textbf{Q1}, \textbf{Q2}, \textbf{Q3} and \textbf{Q4}) by applying distinct prompting techniques (i.e., zero-shot, few-shot, and the two versions of chain-of-thought prompting) on each of the 108 questions in each batch. In particular, each prompt was tested 
on 24 LLMs of varying sizes and configurations, to provide a comprehensive analysis across different architectures and scales. Specifically, the models used in our evaluation are the following:
\begin{itemize}
\item \textbf{Gemma Series} \cite{team2024gemma}: This includes the \texttt{gemma-1.1} models with 2B and 7B parameters, and the more recent \texttt{recurrentgemma} models \cite{de2024griffin} with 2B and 9B parameters, pretrained on 6T tokens of varying sources.
\item \textbf{StableLM Series} \cite{bellagente2024stable}: This includes \texttt{stablelm-2} models with 1.6B and 12B parameters, pretrained on 2T tokens of diverse multilingual and code datasets.
\item \textbf{Qwen Series} \cite{qwen}: This series ranges from 1.8B to 72B parameters, pretrained on 3T tokens of diverse texts and codes, optimized for multilingual chat applications.
\item \textbf{Phi Series} \cite{abdin2024phi}: This includes \texttt{Phi-3} models with parameter sizes of 3.8B (mini), 7B (small), and 14B (medium), pretrained on 3.3T tokens of varying sources and designed for long-context instruction following.
\item \textbf{Yi Series} \cite{young2024yi}: This includes \texttt{Yi-1.5} models with 6B, 9B, and 34B parameters, pretrained on a high-quality corpus of 500B tokens.  
\item \textbf{Llama Series} \cite{touvron2023llama,llama3modelcard}: This includes \texttt{Llama-2} models with 7B and 13B parameters pretrained on 2T tokens of data from publicly available sources, and the more recent \texttt{Meta-Llama-3} models with 8B and 70B parameters, pretrained on over 15T tokens of data from publicly available sources.
\item \textbf{GPT Series} \cite{achiam2023gpt}: This includes OpenAI's \texttt{GPT-3.5} and \texttt{GPT-4} closed-source models, noted for their advanced conversational abilities.
\end{itemize}

\subsection{Evaluation Settings}
\label{sec:es}
To evaluate their reasoning capabilities, we counted the \textit{proportion
of correct answers} by each model. As our questions are binary (i.e., either ``yes'' or ``no''), we expect a model to succeed in at least 50\% of the questions, even if just by chance. 
On the other hand, the question dataset is not fully balanced: 55 out of the 108 test questions have answered ``no''. For this reason, we compare the
results against a \textbf{baseline} (first row of Tables~\ref{tab:results} and~\ref{tab:results:cot}) from a model that always answers ``no'', with an \textit{accuracy} of $55/108\approx51\%$.
The experiments were run thrice on each model with different seeds; the numbers show the \textit{mean accuracy}.
The \textbf{ALL} column provides the \textit{overall accuracy mean} of the models over the concatenation of the four batches, together with its \textit{standard deviation} (from the three runs).
Considering answer correctness as a random event, we estimate the accuracy mean 
through a 95\% \textit{Confidence Interval} (\textbf{CI}) and evaluate the differences between models using standard statistical methods, 
through the normal approximation due to the law of large numbers~\cite{dekking2005modern}. 
As detailed in Section \ref{sec:modeling}, the results in Table~\ref{tab:results} refer to \textbf{Zero-Shot} and \textbf{Few-Shot} prompting. In the zero-shot setting, the model is asked to provide an answer without any additional examples. In the few-shot setting, the model is given a few examples along with the prompt (see Section \ref{sec:fs}).

The results in Table~\ref{tab:results:cot} refer instead to \textbf{CoT} prompting, where the system is asked to provide a
rationale for its answer \textbf{before} and \textbf{after} giving
the answer, respectively. In this case, due to resource constraints,
the two models with 70B parameters were not included in the
evaluation.

\begin{table*}[!ht]
  \caption{Summary of results on 22 LLMs. Columns show average accuracy over 3 runs divided by batch and globally, and the width of a 95\% CI in CoT prompting, before and after providing the answer. The ALL column represents the concatenation of the four batches Q1--Q4, and includes the standard deviation computed over the three runs with different random seeds.}
  \label{tab:results:cot}
\resizebox{\textwidth}{!}{
\begin{tabular}{@{}lcccccc|cccccc@{}}
\toprule
 & \multicolumn{6}{c|}{\textbf{CoT$_{before}$}} & \multicolumn{6}{c}{\textbf{CoT$_{after}$}} \\
\textbf{Model} & \textbf{Q1} & \textbf{Q2} & \textbf{Q3} & \textbf{Q4} & \textbf{ALL} & \textbf{95\% CI} & \textbf{Q1} & \textbf{Q2} & \textbf{Q3} & \textbf{Q4} & \textbf{ALL} & \textbf{95\% CI} \\
\midrule
\textbf{baseline} & 50.93 & 50.93 & 50.93 & 50.93 & 50.93 & $\pm 0$ & 50.93 & 50.93 & 50.93 & 50.93 & 50.93 & $\pm 0$ \\
\midrule
\textbf{gemma-1.1-2b-it} & 57.72 & 54.94 & 50.93 & 60.80 & $ 56.10\pm 1.70$ & $\pm$2.70 & 50.31 & 51.85 & 49.38 & 49.07 & $50.15\pm 0.66 $ & $\pm$2.72\\ 
\textbf{stablelm-2-1\_6b-chat} & 20.99 & 21.30 & 21.30 & 26.23 & $22.45\pm 2.60 $ & $\pm$2.27 & 59.88 & 58.02 & 66.67 & 47.84 & $ 58.10\pm 1.98 $ & $\pm$2.69\\ 
\textbf{Qwen1.5-1.8B-Chat} & 49.07 & 51.23 & 50.31 & 48.46 & $49.77\pm 0.76 $ & $\pm$2.72 & 49.07 & 49.69 & 50.00 & 42.28 & $ 47.76\pm 1.40 $ & $\pm$2.72\\ 
\textbf{recurrentgemma-2b-it} & 47.22 & 47.53 & 50.31 & 48.46 & $48.38\pm 0.95 $ & $\pm$2.72 & 66.98 & 64.20 & 65.12 & 70.68 & $ 66.74\pm 0.27 $ & $\pm$2.57\\ 
\textbf{Phi-3-mini-128k-instruct-3.8B} & 50.62 & 47.84 & 45.68 & 49.07 & $48.30\pm 2.75 $ & $\pm$2.72 & 55.86 & 54.32 & 52.47 & 63.27 & $ 56.48\pm 0.80 $ & $\pm$2.70\\ 
\textbf{Qwen1.5-4B-Chat} & 52.47 & 53.70 & 48.77 & 50.00 & $51.23\pm 2.77$ & $\pm$2.72 & 58.64 & 58.95 & 54.94 & 58.64 & $ 57.79\pm 1.19 $ & $\pm$2.69\\ 
\midrule
\textbf{Yi-1.5-6B-Chat} & 38.27 & 44.44 & 45.99 & 34.57 & $ 40.82\pm 0.72 $ & $\pm$2.68 & 53.40 & 54.94 & 50.00 & 47.84 & $51.54\pm 0.13 $ & $\pm$2.72\\ 
\textbf{Llama-2-7b-chat-hf} & 21.05 & 23.53 & 22.66 & 19.75 & $21.75\pm 0.38 $ & $\pm$2.25 & 61.11 & 58.02 & 58.64 & 66.98 & $61.19\pm 0.94 $ & $\pm$2.65\\ 
\textbf{Meta-Llama-3-8B-Instruct} & 54.63 & 51.85 & 56.17 & 69.75 & $ 58.10\pm 1.68 $ & $\pm$2.69 & 61.11 & 58.95 & 55.56 & 66.36 & $ 60.49\pm 2.15 $ & $\pm$2.66\\ 
\textbf{gemma-1.1-7b-it} & 55.25 & 54.01 & 51.23 & 54.63 & $53.78\pm 0.44$ & $\pm$2.71 & 47.53 & 48.15 & 46.91 & 54.01 & $ 49.15\pm 0.35 $ & $\pm$2.72\\ 
\textbf{Qwen1.5-7B-Chat} & 56.48 & 56.79 & 52.16 & 65.74 & $ 57.79\pm 0.29 $ & $\pm$2.69 & 50.62 & 52.47 & 51.54 & 60.80 & $ 53.86\pm 0.74 $ & $\pm$2.71\\ 
\textbf{Yi-1.5-9B-Chat} & 50.31 & 47.53 & 44.44 & 45.99 & $47.07\pm 0.11$ & $\pm$2.72 & 57.10 & 58.33 & 62.96 & 54.01 & $ 58.10\pm 1.22 $ & $\pm$2.69\\ 
\textbf{recurrentgemma-9b-it} & 57.10 & 55.86 & 50.00 & 53.40 & $54.09\pm 0.44 $ & $\pm$2.71 & 58.95 & 59.57 & 53.40 & 60.49 & $ 58.10\pm 0.75 $ & $\pm$2.69\\
\textbf{Phi-3-small-128k-instruct} & 52.16 & 60.80 & 58.64 & 43.52 & $53.78\pm 0.97 $ & $\pm$2.71 & 65.74 & 70.68 & 63.58 & 75.31 & $ 68.83\pm 0.53 $ & $\pm$2.52\\ 
\midrule
\textbf{stablelm-2-12b-chat} & 19.44 & 21.91 & 22.22 & 20.06 & $ 20.91\pm 0.79 $ & $\pm$2.21 & 45.37 & 50.00 & 46.91 & 54.94 & $ 49.31\pm 0.23 $ & $\pm$2.72\\ 
\textbf{Llama-2-13b-chat-hf} & 30.56 & 34.88 & 23.46 & 24.07 & $ 28.24\pm 1.36 $ & $\pm$2.45 & 50.93 & 50.93 & 50.93 & 52.16 & $ 51.23\pm 0.13 $ & $\pm$2.72\\ 
\textbf{Qwen1.5-14B-Chat} & 50.00 & 52.78 & 49.38 & 54.01 & $ 51.54\pm 0.39 $ & $\pm$2.72 & 55.25 & 58.95 & 54.32 & 66.98 & $ 58.87\pm 0.48 $ & $\pm$2.68\\ 
\textbf{Phi-3-medium-128k-instruct} & 55.86 & 57.72 & 55.86 & 61.11 & $ 57.64\pm 0.82$ & $\pm$2.69 & 62.65 & 64.51 & 61.42 & 67.90 & $64.12\pm 0.61 $ & $\pm$2.61\\ 
\midrule
\textbf{Yi-1.5-34B-Chat} & 57.72 & 59.57 & 59.88 & 57.41 & $ 58.64\pm 0.39 $ & $\pm$2.68 & 53.70 & 58.33 & 52.47 & 47.53 & $ 53.01\pm 0.40 $ & $\pm$2.72\\ 
\textbf{Qwen1.5-32B-Chat} & 56.79 & 59.88 & 59.88 & 60.80 & $59.34\pm 0.29 $ & $\pm$2.67 & 52.78 & 51.54 & 52.78 & 60.19 & $54.32\pm 0.27 $ & $\pm$2.71\\ 
\midrule
\textbf{GPT-3.5-turbo-0613} & 46.30 & 46.30 & 43.21 & 47.53 & $45.83\pm 0.68$ & $\pm$2.71 & 48.15 & 48.77 & 48.46 & 48.15 & $ 48.38\pm 1.20 $ & $\pm$2.72\\ 
\textbf{GPT-4-0613} & 75.93 & 75.93 & 73.46 & 74.38 & $74.92\pm 1.09$ & $\pm$2.36 & 74.69 & 75.31 & 70.68 & 70.99 & $72.92\pm 0.40 $ & $\pm$2.42\\ 
\bottomrule
\end{tabular}
}
\end{table*}

\section{Results}
\label{sec:results}
Table~\ref{tab:results} shows the \textit{average accuracy} results separated by batch for both evaluation settings. 

Concerning \textit{zero-shot prompting}, the most accurate model is GPT-4 with an overall accuracy of 72.30\%, but its difference with
Qwen72B, Llama70B, Phi-small, recurrentgemma-2b are not statistically significant (their confidence intervals
intersect). The other models are mostly statistically indistinguishable from the baseline.
StableLM models are well below even random guesses, but the results are biased by the model
not always providing a Boolean answer, and hence hiding potentially correct but voided answers.\footnote{Around 30\% of the StableLM series model responses were considered void because they did not follow the prompt instructions. These responses were marked as incorrect. In all our experiments, this issue was observed with StableLM models only.}
%
The interbatch analysis shows almost no significant difference, except for Phi-medium being significantly 
better at batch Q4 than at batch Q3, Qwen32B being better at batch Q4 than all others, and Qwen72B being 
significantly worse at batch Q3 than all others. 

Regarding \textit{few-shot prompting}, the results follow the same previous trend, but models perform \emph{worse} in general. The only exception is StableLM, where no void answers were now encountered, and the accuracy is accordingly improved. 
Recall that answers that did not start with ``yes'' or ``no'' were 
labelled as void are considered wrong.
The only models statistically distinguishable from the baseline are
GPT-4, Qwen72B, Llama70B, and Phi-small, all of which have intersecting CIs. Such intersection implies that their performance
is not statistically distinguishable.
The interbatch analysis provides the same outcomes of significant differences
as in one-shot prompting.

The picture is more involved in \textit{CoT prompting}. When the rationale was requested \textbf{before} the answer, GPT-4 was still the best performing model, but most other models showed very
low accuracy (reaching 21\% in StableLM and Llama models). This is
in part explained by the surge of void answers, caused by the models
not ending their response with a Boolean statement (specifically stablelm-2-1\_6b-chat, Phi-3-mini-128k-instruct-3.8B, Llama-2-7b-chat-hf, Yi-1.5-9B-Chat, Phi-3-small-128k-instruct, and stablelm-2-12b-chat have around 20\% of void answers). Yet even 
accounting for these void answers, the performance degraded greatly
in this case.
Interestingly, when the rationale is given \textbf{after} the Boolean answer, the performance of most models
tended to increase (in most cases, significantly), even though the
beginning of the output string is expected to be unchanged.

\subsection{Invariability}

Recall that the first three batches of test prompts contain exactly the same 
questions with very small variations of context and terminology, and the fourth 
batch only negates the question (and its answer) from the first one. Hence, if 
a model is invariable w.r.t.\ language modifications (i.e., remains stable
under syntactic variations of the same semantic element), it should provide
the same answers in each batch. Invariability is independent of
accuracy or correctness: when a model is invariable, it should provide the
same answer to the same question, regardless of whether the answer is correct
or not. More precisely, when measuring invariability, our goal is to
measure how often the answers of the models change depending on the
prompt.

To find out whether a model is invariable in this sense, it is not enough to check the 
proportion of correct answers it provides on each batch (accuracy), but one must verify the 
differences in the answers:
importantly, two models may have 50\% accuracy
each, and do not coincide in any answer; thus showcasing very
variable behaviour. This means that we must 
analyse how often the answers coincide.
We check the pairwise differences in behaviour of the 
same model between batches using a \textit{McNemar test}~\cite{McNe47}, 
a statistical evaluation designed explicitly
for this purpose. Very briefly, the test counts how unbalanced the
discordants (the different answers) are in one direction or the other,
hence quantifying the stability of answers. 

We applied the McNemar test only to the pairs that were not already deemed as
statistically different from the accuracy analysis; all other cases clearly yield variable answers. All the pair-wise statistics 
obtained from the McNemar test
yield a $p$-value over 0.1, which in a nutshell means
that the differences in answers between batches (within any given model) are not 
statistically significant. We have thus no evidence to believe that the models
behave differently for different batches of questions, or in our terms, we cannot say that the behaviour varies. This result is interesting, as it
suggests that prompt engineering has limited impact in these cases, at
least up to the point of the linguistic variants tested.

We also observed an invariability \emph{between} models; that is, the answers from different
models were also similar in general. This observation is important, because it 
implies that despite their architectural, size, and training differences,
when it comes down to the logical questions in our dataset, all models are more or less 
equal (and \emph{none} very proficient). This insight thus allows us to study ways of circumventing the
problem for \emph{all} models, without the need for \emph{ad-hoc} solutions
for each of them. This is a potential area for future research.

\section{Conclusions and Future Work}

In this article, we provided an in-depth analysis of the reasoning abilities of Large Language Models (LLMs), 
by presenting a new dataset of questions for logical reasoning which turned out to be challenging for 24 different LLMs in distinct prompt engineering settings. In our experiments, most LLMs behaved in a similar fashion 
in both a zero-shot and a 
few-shot setting, 
except that the performance of most models
degraded 
for few-shot reasoning.
Through a McNemar test, we also verified that the behaviour of the models is mainly invariant w.r.t.\ small language and stylistic variations (with only a 
few exceptions).
The answers also tended to be invariant w.r.t. the chosen
model. By understanding the cases where 
LLMs fail, we hope 
to circumvent the problem in future models, to be more robust and effective than fine-tuning or filtering.

This work points to promising research directions, especially in misinformation detection \cite{zhou2020survey}. With the overwhelming volume of online content, human fact-checking is inadequate. Pseudo-automatic knowledge-based systems can aid but struggle with the labor-intensive task of updating knowledge bases \cite{pan2018content,whitehouse2022evaluation}. While LLMs could be beneficial, they often fail to effectively convert natural language into structured knowledge \cite{viviani2017credibility,allen2023knowledge}. Therefore, this research area remains open for further exploration.

\section*{Acknowledgment}
We acknowledge the CINECA award under the ISCRA initiative, for the availability of high-performance computing resources and support. This work was partially funded by the European Union – Next Generation EU within the project NRPP M4C2, Investment 1.,3 DD. 341 - 15 march 2022 – FAIR – Future Artificial Intelligence Research – Spoke 4 - PE00000013 - D53C22002380006
and by the MUR under the grant ``Dipartimenti di Eccellenza 2023-2027'' of the Department of Informatics, Systems and Communication of the University of Milano-Bicocca, Italy.

\bibliographystyle{IEEEtran}
\bibliography{main}

\end{document}